\documentclass[11pt,a4paper,onecolumn,oneside]{article}
\usepackage[T1]{fontenc}
\usepackage[utf8]{inputenc}
\usepackage{lmodern}
\usepackage[a4paper,margin=25mm]{geometry}
\usepackage{microtype}
\usepackage{amsmath,amssymb}
\usepackage{array,booktabs,tabularx,longtable}
\usepackage{float,flafter,placeins,graphicx}
\usepackage{tikz}
\usetikzlibrary{arrows.meta,positioning,calc}
\usepackage{setspace}
\usepackage[font=small,labelfont=bf,skip=5pt]{caption}
\usepackage[numbers,sort&compress]{natbib}
\usepackage{xurl}
\usepackage[hidelinks]{hyperref}
\newcommand{\PreprintAuthor}[2]{#1\textsuperscript{#2}}
\newcommand{\PreprintAffiliation}[2]{\textsuperscript{#1}#2\par}
\newcommand{\PreprintFrontMatter}[4]{%
  \begin{center}
    {\LARGE\bfseries #1\par}\vspace{.8em}
    {\large #2\par}\vspace{.55em}
    {\small #3}\vspace{.4em}
    {\small\textsuperscript{*}Corresponding author: #4\par}
  \end{center}
}
\newcommand{\PreprintKeywords}[1]{\noindent\textbf{Keywords:} #1\par}
\newcommand{\PreprintBibliography}[1]{%
  \begingroup\raggedright
  \bibliographystyle{unsrtnat}%
  \bibliography{#1}%
  \endgroup
}

\hypersetup{
  pdftitle={On the Complete Feature Set for ADHD Classification in Children and Adolescents Based on EEG Signals},
  pdfauthor={Javier Sanchis, Álvaro Navarro, Jorge García-Carrasco, Tahar Kechadi, Juan Trujillo}
}
\begin{document}
\PreprintFrontMatter
  {On the Complete Feature Set for ADHD Classification in Children and Adolescents Based on EEG Signals}
  {\PreprintAuthor{Javier~Sanchis}{a,*},
   \PreprintAuthor{{\'A}lvaro~Navarro}{a},
   \PreprintAuthor{Jorge~Garc{\'i}a-Carrasco}{a},
   \PreprintAuthor{Tahar~Kechadi}{b},
   \PreprintAuthor{Juan~Trujillo}{a}}
  {\PreprintAffiliation{a}{Lucentia Research, Department of Software and Computing Systems, University of Alicante, Carretera San Vicente del Raspeig s/n, Sant Vicent del Raspeig, 03690, Alicante, Spain}
   \PreprintAffiliation{b}{School of Computer Science and Informatics, University College Dublin, Belfield, Dublin 4, D04 V1W8, Ireland}}
  {Javier Sanchis, \href{mailto:javier.sanchis@ua.es}{javier.sanchis@ua.es}}
\begin{abstract}
Attention-deficit/hyperactivity disorder is a neurodevelopmental condition that
can affect attention, activity and impulse control. Electroencephalography (EEG)
records brain activity as children and adolescents perform cognitive tasks.
However, comparing features derived from these recordings requires consistent
signal preparation, shared analysis windows and documented quality decisions.
We curated 669 existing BALLADEER recordings from 156 participants across five
task conditions so that researchers can compare seven EEG feature families
on the same time intervals for each participant and task. The dataset links
continuous signals to precomputed measurements from 115 feature definitions or
templates and records the quality decisions made for each recording. Analyses
must still account for repeated measurements within participants, residual
artefacts and differences between acquisition systems.
Technical validation covered signal integrity, numerical checks on feature
estimators and completeness of the released files. In every task, the median
proportion of recording time retained after preprocessing was 100\%.
To illustrate reuse, we report saved participant-held-out evaluations comparing
XGBoost using all seven families on eight-second windows with a predictor based only on recruitment-group
prevalence in the training fold. Recruitment-group discrimination was stronger
in Robots than in the three Slackline conditions. These descriptive results
from overlapping cohorts do not establish diagnostic validity.
The dataset, curation software and aggregate results for the predictive example
are publicly available.
\end{abstract}
\PreprintKeywords{EEG, ADHD, Signal preprocessing, Feature extraction, Quality control, Research data}

\section{Introduction}

Attention Deficit Hyperactivity Disorder (ADHD) is a neurodevelopmental condition
characterised by persistent inattention and/or hyperactivity-impulsivity that
interferes with everyday functioning. Its heterogeneous presentation motivates
research into complementary measurements of the processes underlying these
behaviours. Clinical diagnosis requires assessment of developmental history,
symptoms and functioning across settings. Electroencephalography (EEG) provides
non-invasive measurements of electrical brain activity that can support this
research, including investigation of candidate neurophysiological markers during
cognitive tasks. These measurements complement the study of ADHD but do not,
by themselves, constitute an established diagnostic test
\cite{nice2018adhd,peterson2024diagnostictools}.

A central approach to analysing EEG is to derive features: numerical descriptions
of properties such as signal amplitude, oscillatory power, irregularity and
relationships between electrodes. These representations make particular signal
properties explicit and provide inputs for statistical analysis, feature selection
and machine learning. Recent EEG--ADHD literature includes spectral, entropy and
connectivity measures, reflecting interest in complementary descriptions of brain
activity \cite{quinterolopez2025review}. Engineered features also complement
approaches that learn representations directly from signals. Retaining continuous
EEG alongside precomputed features therefore enables researchers to investigate
both approaches and examine how representation choices affect downstream findings.

Public datasets provide an essential foundation for such work. Existing resources
include paediatric attention-task recordings and broader transdiagnostic
collections such as the Healthy Brain Network
\cite{nasrabadi2020eegdata,alexander2017hbn}. The BALLADEER dataset extends this
landscape with EEG recorded during cognitive activities alongside behavioural,
cognitive and other physiological measurements. Its EEG recordings cover CogniFit,
Attention Robots and selected levels of Attention Slackline, providing
opportunities to investigate representations across different recording contexts
\cite{trujillo2026balladeer}.

However, access to recordings does not itself provide the aligned signals,
feature definitions and quality information needed to compare representations
on matched data. In BALLADEER, reuse requires accounting for acquisition-system
differences, recording discontinuities and channel problems before defining
analysis windows and computing features. When studies intended to compare
representations use different preparation procedures or recording intervals,
their findings can also reflect these choices. Automated EEG pipelines already
demonstrate the value of documented preprocessing and feature extraction
\cite{gilavila2023discover}. For these task recordings, the remaining need is a
reusable collection that links the resulting measurements to their source
intervals, curation decisions and calculation states.

To address these reuse barriers, we prepared a secondary curated resource
containing 669 eligible recordings from 156 children and adolescents across five
task conditions: CogniFit, Robots and selected levels of Slackline. The resource
retains continuous EEG alongside complementary feature tables and source-linked
quality records. It uses established processing methods without recruiting a
new cohort or treating the different acquisition systems as equivalent. Its
contribution is the additional curation information, aligned measurements and
documented computational definitions made available for reuse.

The resource makes four contributions:
\begin{itemize}
\item \textbf{Auditable signal derivatives.} Label-blind eligibility decisions,
expert quality assessments and processing records accompany curated continuous
EEG, allowing researchers to inspect how recordings were selected and transformed.
\item \textbf{Aligned multi-family representations.} Seven feature families
encompass 115 definitions or templates computed on shared analysis intervals.
Explicit calculation states distinguish unavailable or undefined measurements,
while catalogues document their indexed expansion across channels and other
positions.
\item \textbf{Documented validation and reproduction resources.} Numerical
reference checks, structural integrity checks, configurations and preparation
software support inspection and reproduction of the released derivatives.
\item \textbf{A participant-held-out baseline for reuse.} Across four overlapping
task cohorts, saved evaluations compare XGBoost using all seven feature families
on eight-second windows with a predictor that assigns every held-out participant
the proportion of ADHD recruits in the training fold. These retrospective results illustrate recruitment-group
discrimination, which was weaker in Slackline than in Robots; they do not
establish diagnostic validity.
\end{itemize}

The resource supports representation comparisons, feature-selection research and
alternative signal analyses, provided that reuse accounts for participant
dependence, residual artefacts and task-device differences.

\section{Materials and Methods}

\begin{figure}[htbp]
\centering
\includegraphics[width=\linewidth]{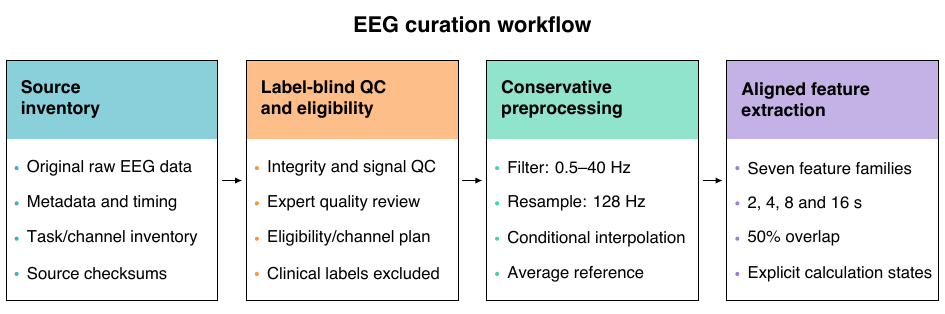}
\caption{EEG curation workflow, from source inventory to aligned feature extraction.}
\label{fig:eeg-curation}
\end{figure}

The curation workflow combines source auditing, conservative preprocessing
selected from technical Quality Control (QC) evidence, and aligned feature
extraction (Figure~\ref{fig:eeg-curation}). Clinical and demographic labels did not determine eligibility,
preprocessing or feature values. The release links the processed signals and
features to their QC and processing records.

\subsection{Source data and cohort}

\subsubsection{Source dataset, participants and ethics}
\label{subsec:source-cohort}

This resource derives from five EEG tasks in the publicly available BALLADEER
dataset \cite{trujillo2026balladeer,trujillo2026balladeerdata}. Pseudonymous
identifiers support linkage to compatible source records. The original
recordings and non-EEG modalities remain in the source dataset.

The source demographic file contains 158 participant records. This resource
includes 669 eligible task recordings from 156 of those participants. In the
full source cohort, participants were aged 6-18 years, with 96 coded as male
and 62 as female. Participants were asked to discontinue ADHD medication for
at least 24 hours before collection \cite{trujillo2026balladeer}.

The field \emph{group} distinguishes 62 participants recruited with a formal
pre-existing ADHD diagnosis from 96 recruited as comparisons. The separate
field \emph{diagnosed} records the clinical team's assessment-derived status
as \emph{yes}, \emph{no}, or \emph{undetermined}. It is not equivalent to
pre-existing diagnosis. All ADHD recruits have status \emph{yes}, whereas
the comparison group contains all three outcomes and should not be treated as
clinically confirmed healthy controls. Table~\ref{tab:cohort-demographics}
cross-classifies these fields, which were excluded from processing decisions.

\begin{table}[ht]
\centering
\caption{Recruitment and assessment-derived label composition.}
\label{tab:cohort-demographics}
\small
\setlength{\tabcolsep}{4pt}
\begin{tabular}{@{}>{\raggedright\arraybackslash}p{0.28\textwidth}rrrr@{}}
\toprule
\textbf{Group}    & \textbf{Group total} &
\multicolumn{3}{c}{\textbf{Diagnosed}} \\

\cmidrule(l){3-5}
                                        &                           & \emph{yes} & \emph{no} & \emph{undetermined} \\
\midrule
ADHD recruitment                 & 62                        & 62 & 0 & 0 \\
Control recruitment                     & 96                        & 26 & 50 & 20 \\
\bottomrule
\end{tabular}
\end{table}

The original study received approval from the University of Alicante Ethics
Committee (UA-2023-05-17\_1) and the ISABIAL Ethics Committee (PI2022-139).
Participants or their legal guardians, as applicable, provided written consent
for participation, collection, and anonymised public sharing
\cite{trujillo2026balladeer}. No new recruitment or acquisition was undertaken.

\subsubsection{Experimental tasks and EEG acquisition}
\label{subsec:tasks-acquisition}

The five task conditions comprised CogniFit, Robots (Attention Robots), and
Slackline L1, L6 and L11, where L denotes the Attention Slackline level.
CogniFit involved adaptive exercises in
perception, attention, coordination, and problem solving. Robots
required participants to select robots matching target criteria in a
$14\times47$ grid, with 20 seconds per row. Slackline comprised five-minute
levels 1, 6, and 11, in which participants pressed a button in response to
on-screen flag patterns to control an avatar's progress. These levels required
matching one target, either of two targets, or a two-target sequence,
respectively
\cite{teruel2024measuring,trujillo2026balladeer}. EEG spanned task onset to
completion, excluding tutorial or instruction intervals. No dedicated
resting-state condition was collected. Task identifiers and acquisition
characteristics are summarised in Table~\ref{tab:source-eeg-tasks}.

Prior behavioural work described and validated Slackline
\cite{teruel2024measuring}. A separate study of Robots reported preliminary
behavioural convergent validity against D2-R \cite{panagiotidis2026robots}.
That study's behavioural sample and outcomes do not establish EEG or diagnostic
validity for the present resource.

\begin{table}[ht]
\centering
\caption{Source EEG tasks and acquisition characteristics.}
\label{tab:source-eeg-tasks}
\small
\setlength{\tabcolsep}{4pt}
\begin{tabular}{@{}>{\raggedright\arraybackslash}p{0.18\textwidth}>{\raggedright\arraybackslash}p{0.09\textwidth}>{\raggedright\arraybackslash}p{0.40\textwidth}>{\raggedright\arraybackslash}p{0.23\textwidth}@{}}
\toprule
Source task & ID & Brief protocol & Acquisition \\
\midrule
CogniFit & BAL01 & Adaptive computerized cognitive exercises & EPOC X/Plus, 128 or 256 Hz \\
Robots & BAL02 & Visual search for robots matching target criteria & EPOC X/Plus, 128 or 256 Hz \\
Slackline L1 & BAL03 & Button response to one on-screen flag target & CGX Quick-32r, 500 Hz \\
Slackline L6 & BAL04 & Button response to either of two on-screen flag targets & CGX Quick-32r, 500 Hz \\
Slackline L11 & BAL05 & Button response to an on-screen two-flag sequence & CGX Quick-32r, 500 Hz \\
\bottomrule
\end{tabular}
\end{table}

CogniFit and Robots used 14-channel Emotiv EPOC X or EPOC Plus at
128 Hz, except seven EPOC Plus recordings in the valid 256-Hz mode. The
Slackline tasks used CGX Quick-32r at 500 Hz, contributing 29 scalp EEG channels. Operators checked
vendor sensor-contact indicators before each session
\cite{trujillo2026balladeer}. Adapters converted amplitudes to volts and
assigned standard 10-20 template coordinates, not participant-specific
digitisations. Emotiv streams included vendor Sinc and notch conditioning,
whereas CGX retained minimal hardware conditioning. No common offline
processing or undocumented acquisition reference was assumed at ingestion.
Task, device, montage, and duration differences preclude treating the tasks as
interchangeable conditions.

\subsubsection{Source data audit and recording eligibility}
\label{subsec:source-audit}

The audit enumerated $158\times5=790$ participant-task combinations, checked
agreement between metadata and directories, located EEG CSV files, and
required readable samples and valid timestamps. Missing tasks were recorded
as unavailable. Relative paths and input checksums preserve source identity.

For $N$ samples with first and last valid timestamps $t_1,t_N$, duration was
$T=t_N-t_1$ and effective sampling frequency was $f_{\mathrm{eff}}=(N-1)/T$.
Within each task, duration outside
$[P_{10}-1.5(P_{90}-P_{10}),\,P_{90}+1.5(P_{90}-P_{10})]$ was flagged.
Sampling frequency failed independently when
$|f_{\mathrm{eff}}-f_{\mathrm{expected}}|/f_{\mathrm{expected}}>0.20$,
using the Emotiv header value or 500 Hz for CGX.

Eligibility outcomes are reported in Section~\ref{subsec:source-inventory-validation}. Absence or failure in one task did not exclude a participant's other recordings. Audit tables retain each measurement and decision reason.

\subsection{Quality characterisation and preprocessing}

Measurements at window, channel, and recording scales informed label-blind
selection and a gap-aware preprocessing sequence.

\subsubsection{Raw EEG quality characterisation}
\label{subsec:raw-qc}

Raw QC combined 4s temporal, 8s spectral, and full-recording measurements
before offline processing. This multiscale approach drew on established EEG
quality frameworks \cite{nolan2010faster,bigdelyshamlo2015prep,jas2017autoreject,
gabarddurnam2018happe,pedroni2019automagic} and reporting recommendations
\cite{pernet2020cobidas}. Measurements were compared within task-acquisition
strata for EPOC X, EPOC Plus, and CGX, accounting for device-dependent spectra
\cite{mikhaylov2024devices}. They indicate technical integrity or possible
contamination, not biological normality \cite{delorme2023alone}.
Table~\ref{tab:raw-qc-overview} lists the retained measurements.

\begin{table}[H]
\centering
\caption{Raw EEG QC measurements retained at each analysis scale.}
\label{tab:raw-qc-overview}
\small
\begingroup
\setstretch{1.1}
\renewcommand{\arraystretch}{1.25}
\setlength{\tabcolsep}{4pt}
\begin{tabular}{@{}>{\raggedright\arraybackslash}p{0.18\textwidth}>{\raggedright\arraybackslash}p{0.74\textwidth}@{}}
\toprule
Analysis scale & QC measurements \\
\midrule
4s temporal windows (50\% overlap) & Finite-sample, Not a Number, and infinity fractions, Standard Deviation, Median Absolute Deviation, median-centred Root Mean Square, peak-to-peak amplitude, maximum absolute derivative, line length, flat-sample and repeated-value fractions, longest flat run, skewness, Fisher excess kurtosis, median-centred zero-crossing rate, median pairwise Spearman channel correlation, and median Global Field Power. \\
8s spectral windows (50\% overlap) & Absolute and relative delta, theta, alpha, beta, and high-beta power, slow-drift ratio, common and extended high-frequency ratios, 50-Hz line-noise ratio and supported harmonics, spectral entropy, Theil-Sen spectral slope, 95\% spectral edge frequency, median pairwise spectral correlation, and median absolute spectral deviation. \\
Full recording & Recorded and usable duration, expected and observed sampling frequency, signed sampling-frequency error, interval jitter, duplicate and backward timestamps, invalid-row count, gap count and duration, continuous-block count, channel inventory, full-recording median channel and nearest-neighbour correlations, and persistent spectral deviation. \\
\bottomrule
\end{tabular}
\endgroup
\end{table}

Both window scales used 50\% overlap, excluded incomplete windows, and never
crossed gaps. Values were summarised within recordings before calibration, so
overlapping windows were not treated as independent observations. The scales
balance temporal localisation and spectral support rather than represent
empirically established optima \cite{welch1967fft,harris1978windows}.

Temporal metrics included Standard Deviation (SD), Median Absolute Deviation
(MAD), and median-centred Root Mean Square (RMS). Flat and repeated-value
fractions used a $10^{-9}$ V change threshold. Median-centred zero crossings
used a $\pm1\,\mu$V dead-band. No temporal metric alone established exclusion.

Spectral QC used Welch Power Spectral Density (PSD) estimates with 4s Hamming
segments, 50\% overlap, constant detrending, mean averaging, and density
scaling. Each 8s window supplied three segments and a 0.25-Hz grid.
Absolute and relative powers in 1--4, 4--8, 8--13, 13--30, and 30--40 Hz
were descriptive and could not independently trigger rejection. Technical
ratios covered 0.5--1, 30--40, and, where supported, 40--100 Hz. The line-noise
ratio compared $50\pm1$ Hz power with adjacent 2-Hz bands.

Recording-scale checks assessed clocks, channel inventory, continuity, and
persistent channel behaviour. A timestamp gap exceeded
$1.5/f_{\mathrm{expected}}$, with missing duration
$G=\sum_i\max(0,\Delta t_i-f_{\mathrm{expected}}^{-1})$.
EPOC observed sampling frequency was the inverse median positive timestamp
interval. For CGX, timestamp span described transport rate, while the sample
counter defined discontinuities and the analysis clock. Gap count, duration,
and block count were retained separately. Every metric included units,
parameters, source segments, and a state of computed, not computable, not
applicable, or invalid input.

\subsubsection{QC interpretation and preprocessing selection}
\label{subsec:qc-interpretation}

Metrics had prespecified roles as integrity failures, robust warnings,
descriptive evidence, or unavailable evidence. Nonfinite input, invalid rows,
duplicate or backward timestamps, and channel-inventory violations used
absolute rules. Warning thresholds used one profile per recording-task rather
than pooled overlapping windows. The calibration subset comprised 60\% of
each task-acquisition stratum, capped at 80 participants.

For calibration median $m$, $s=1.4826\,\mathrm{MAD}$, and quantile $Q$, the
direction-specific limits were
\[
U=\max\{Q_{0.99},m+6s\},\qquad L=\min\{Q_{0.01},m-6s\}.
\]
The MAD scaling constant provides Gaussian consistency without requiring
normally distributed metrics \cite{rousseeuw2018anomaly,kovacevic2025differences}.
These conservative limits flagged technical extremes. Biology-sensitive
metrics, including band power and spectral edge, could not independently
trigger rejection or preprocessing. Clinical, demographic, and downstream
outcome variables were unavailable during calibration.

We assessed persistence using the proportion of affected windows and their
combined non-overlapping duration. A recording warning required at least 5\% affected
channel-window evidence. Interpolation recommendations required at least three
warning metrics spanning three independent mechanisms
\cite{bigdelyshamlo2015prep,jas2017autoreject}. Automatic states were
\emph{pass}, \emph{pass with warnings}, or \emph{review}. Recommendations
alone did not alter samples.

Experts checked the signal quality of 103 selected recording-task pairs without
access to clinical labels, examining signal traces, PSD, timing integrity,
spatial patterns, and 4s/8s quality measurements. Their assessments helped
determine which recordings to exclude and which channels to interpolate, while
the calibrated automatic thresholds remained unchanged. The assessments,
decisions, and reasons are documented in the released QC records.

Unresolved review states were retained with warnings under the conservative
policy. Final exclusion counts are reported in Section~\ref{subsec:raw-qc-decision-audit}.

\subsubsection{Preprocessing and derivative generation}
\label{subsec:preprocessing}

The same preprocessing operations were applied to EPOC and CGX recordings,
with acquisition-specific QC warning thresholds. Continuous blocks were processed
independently using a fourth-order zero-phase Butterworth second-order-section
filter at 0.5--40 Hz \cite{widmann2015filtering}. Polyphase anti-alias
resampling converted 256-Hz and 500-Hz input to 128 Hz. Native 128-Hz
recordings retained their sampling rate.

Blocks no longer than 1 s could be omitted only when cumulative omission did
not exceed 0.1\% of recording duration. Approved channels were interpolated
using inverse-distance weighting from the four nearest valid template
electrodes, capped at 10\% of the montage. Average reference followed
interpolation. Every omission and channel action was logged.

Each eligible recording yielded one MNE-FIF signal file with channel order,
montage information, boundary annotations, and records linking it to the source
and applied processing. Postprocessing QC repeated compatible raw measurements to assess
retained duration, channel actions, continuity, and expected temporal,
spectral, and connectivity changes. Results are reported in
Section~\ref{subsec:postprocessing-validation}.

\subsection{Feature extraction}
\label{subsec:feature-framework}

All families used the same preprocessing derivatives and deterministic epoch
index, with verified sampling frequency, channel order, blocks, and window
identities. Extractors could not alter samples or epoch bounds. Each feature
had a stable identifier, unit, scope, applicability rule, parameters,
implementation reference, and invalid-value policy. Undefined results remained
reason-coded nulls. Imputation, scaling, feature selection, dimensionality
reduction, and modelling remain downstream operations.

The 115 definitions or templates expand across channels and other indexed
positions rather than forming 115 predictor columns. Table~\ref{tab:feature-family-framework}
summarises definition counts and physical row representations;
Figure~\ref{fig:family-support} displays configured duration support.

\begin{table}[H]
\centering
\caption{Scientific feature definitions and table organisation.}
\label{tab:feature-family-framework}
\footnotesize
\begingroup
\setstretch{1.05}
\renewcommand{\arraystretch}{1.2}
\setlength{\tabcolsep}{2.5pt}
\begin{tabular}{@{}>{\raggedright\arraybackslash}p{0.21\textwidth}>{\raggedright\arraybackslash}p{0.28\textwidth}>{\raggedright\arraybackslash}p{0.32\textwidth}>{\raggedleft\arraybackslash}p{0.10\textwidth}@{}}
\toprule
Family & Each row represents & Row multiplicity and duration support & Count \\
\midrule
Temporal and statistical & One channel and epoch & One per channel at all four durations & 17 \\
Conventional spectral & One channel and epoch & One per channel at all four durations & 18 \\
Aperiodic spectral & One channel-epoch summary or one periodic peak & None at 2s and 4s, one per channel at 8s and 16s, with variable peak rows & 18 \\
Complexity and fractal & One channel and epoch & One per channel. Four definitions null at 2s, one at 4s & 11 \\
Functional connectivity & Epoch-connectivity matrix with packed channel pairs & 1 for 2s and 4s, and 26 for 8s and 16s & 26 \\
Graph and network & Epoch-graph representation with global and nodal metrics & 3 for 2s and 4s, and 28 for 8s and 16s & 15 \\
Time-frequency decomposition & One channel, epoch, and wavelet & Three per channel at all four durations, 61 values per row & 10 \\
\midrule
\textbf{Total} & & & \textbf{115} \\
\bottomrule
\end{tabular}
\endgroup
\end{table}

\subsubsection{Analysis units and epoch construction}
\label{subsec:epoching}

Shared windows had durations $d\in\{2,4,8,16\}$ s and overlap $o=0.5$.
Window length $W=df_s$ and step $S=d(1-o)f_s$ had to be positive integer
sample counts. At 128 Hz, lengths were 256, 512, 1,024, and 2,048 samples,
with steps half as long. Bounds were half-open intervals.

\begin{figure}[htbp]
\centering
\includegraphics[width=\linewidth]{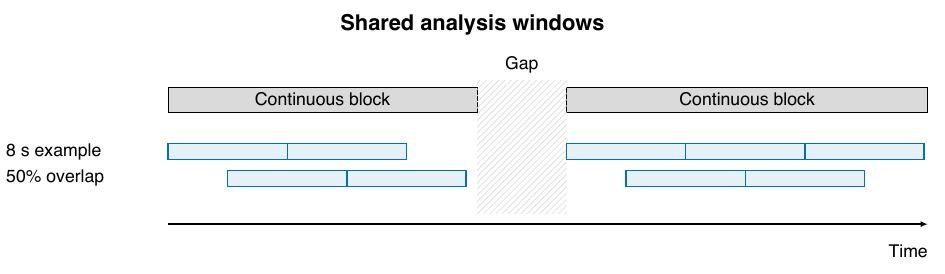}
\caption{Shared analysis windows within continuous recording blocks.}
\label{fig:shared-windows}
\end{figure}

Each window lay within one validated continuous block and could not overlap a
positive-duration exclusion annotation. Incomplete tails were counted and
discarded without padding, reflection, or joining across gaps. The active
epoch recipe left preprocessed samples, channels, and eligibility unchanged.
The windowing example in Figure~\ref{fig:shared-windows} is schematic, not a
measured EEG recording.

Each interval retained participant, recording, task, device, channel-schema,
block, and configuration identifiers. The \emph{window\_id} identified the
derivative, recording, block, bounds, and configuration. The separate
\emph{analysis\_interval\_id} identified the subject, task, and source-anchored
time interval independently of feature family or channel layout. These keys
support aligned comparisons and future linkage to compatible modalities.
Epochs, channels, and recordings from one participant remain dependent.

\subsubsection{Feature families and representations}
\label{subsubsec:feature-families}

Temporal and statistical features describe amplitude, dispersion, shape and
variation, including Interquartile Range (IQR), line-length rate and Hjorth
parameters \cite{quinterolopez2025review,alqahtani2026deep}. Conventional spectral
features summarise Welch-derived powers, ratios, dominant frequency, entropy and
spectral edges; ratios accompany their constituent powers and are not standalone
diagnostic markers
\cite{quinterolopez2025review,dakwarkawar2024aperiodic}. Aperiodic spectral features
separate background shape and peaks \cite{donoghue2020parameterizing,gerster2022separating}.
The recorded FOOOF 1.1.1 backend fits a fixed no-knee 1--30 Hz model at 8s/16s,
yielding 16 candidate predictors and two diagnostics including Mean Absolute
Error (MAE), with a separate variable-length peak inventory.
These hypothesis-driven descriptors have cohort-dependent ADHD associations
\cite{dakwarkawar2024aperiodic,chen2024aperiodic,panda2026relating}.

Complexity and fractal features describe irregularity, finite-scale dependence,
pattern diversity and fractal structure; these measures are not interchangeable
\cite{lau2022complexity,quinterolopez2025review}. Time-frequency decomposition uses
four-level Discrete Wavelet Transforms (DWTs) with three mother wavelets to describe
carrier activity and amplitude modulation across temporal scales
\cite{mallat1989wavelet,safi2026amplitude}. Both families are exploratory.

Functional connectivity measures signed broadband Pearson correlation and five
spectral estimators in five bands; reference, volume conduction, geometry and
estimator choice affect sensor-space interpretation \cite{bastos2016connectivity}.
Graph and network features use the connectivity matrices to describe full weighted topology without data-fitted
thresholds \cite{rubinov2010network}: 15 metrics (five nodal, ten global) use 28
representations from three signed-correlation transforms and direct spectral weights.
They reflect construction choices as well as neural interactions.

\begin{figure}[htbp]
\centering
\includegraphics[width=\linewidth]{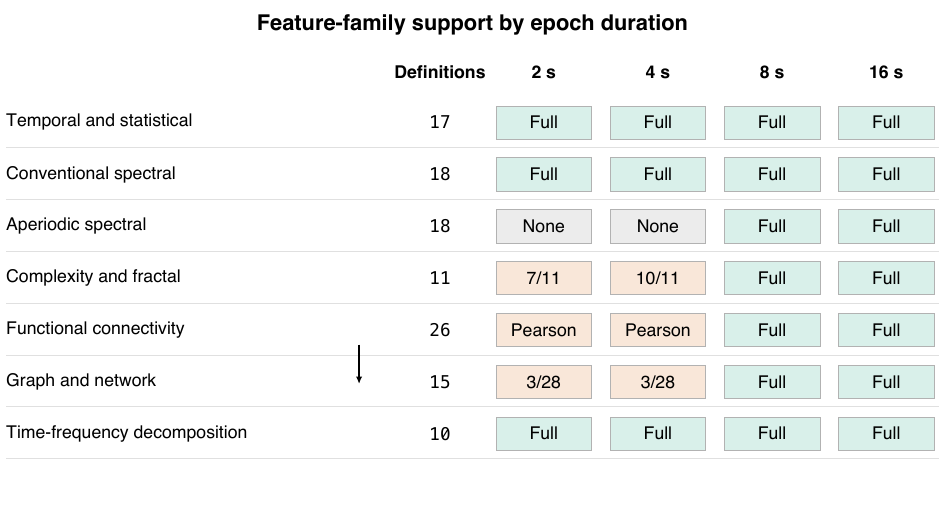}
\caption{Configured feature-family support by epoch duration.}
\label{fig:family-support}
\end{figure}

At 2s/4s, windows support subsets of the complexity features,
Pearson-only connectivity and three graph representations; spectral connectivity
requires at least seven Welch segments at 8s/16s
(Figure~\ref{fig:family-support}). Configured support does not guarantee valid values: undefined results
retain reason-coded nulls. Appendix~\ref{app:feature-definitions} specifies exact
estimators and fixed conventions.

\subsubsection{Metadata, provenance and reproducibility}
\label{subsubsec:metadata-provenance}

Stable identifiers link source recordings, preprocessed signals, continuous
blocks, windows and feature definitions. Processing settings, calculation states
and validation outcomes accompany these links. Automated checks verified source
identity, table structure, window alignment and graph-to-connectivity
correspondence. The companion software supplies the configurations and
computational environment for reproduction; Section~\ref{sec:repository-release-layout}
describes the reuse documentation.

\subsection{Technical validation procedures}
\label{subsec:validation-procedures}

Technical validation examined whether the released derivatives were complete,
traceable, internally consistent, and suitable for reuse. It covered source-file
eligibility, raw-signal QC decisions, paired preprocessing checks, derivative
integrity, and the integrated feature inventory. These checks establish the
technical properties of the resource. They do not establish clinical validity
or the predictive value of any feature.

Recording-level 8s QC summaries were compared using paired raw and preprocessed medians. Sample-count, sampling-frequency, block-correspondence, finite-sample and channel-inventory checks assessed derivative integrity. Saved Robots (BAL02) reference checks used known signals, synthetic spectra, and graph or wavelet invariants. Numerical checks assessed implemented estimators, while a separate inventory audit assessed task-family coverage, file completeness and lineage. Outcomes are reported in Section~\ref{sec:technical-validation}.

\subsection{Predictive utility demonstration}
\label{subsec:predictive-utility-methods}

We report saved participant-held-out evaluations distributed with version 1.1.0
of the preparation software, alongside the code for the predictive example
\cite{sanchis2026eegsoftware110}.
The resource
covers five conditions; this example covers four overlapping cohorts: Robots
and Slackline L1, L6 and L11. The endpoint was recruitment group, with ADHD
recruitment positive, not assessment-derived diagnosis. Each task used all-handcrafted (ALL)
features from the seven families, XGBoost and 8s windows; tuning and fitted
parameters remained task- and fold-specific.

The original outcome-stratified evaluation used three repetitions of five outer
participant folds and four inner folds. All epochs from a participant stayed
together. Training-participant summaries supported availability filtering,
median imputation and robust scaling. Correlation reduction and family-balanced
F ranking retained up to 64 scientific inputs plus eligible missingness indicators.
Preparation, feature selection and transformations were fitted within inner-training
folds. Candidate transformations were identity or Yeo--Johnson. Epoch training
weights accounted for participant epoch counts before class weighting. Mean
participant area under the receiver operating characteristic curve (AUROC)
across inner-validation folds determined candidate selection. We then refitted
the selected preparation and classifier on outer-training data.

An unweighted logistic sigmoid with intercept and inverse regularisation
strength $C=10^6$ was fitted to the selected candidate's pooled inner-validation
participant-mean raw scores. After outer-training refitting, this calibration
function transformed held-out epoch scores \emph{before} averaging probabilities
within participant. Thus calibration was fitted to participant-mean scores but
applied to epoch scores. Decisions used probability 0.5. The comparator used
matching within-task partitions and predicted outer-training recruitment
prevalence without EEG inputs.

Saved metrics pool the five outer test folds within each repetition, then average
the three repetition metrics. Stored 95\% intervals use 1,000 outcome-stratified
participant-cluster resamples retaining each person's three predictions. They
condition on fixed predictions, excluding refitting and selection uncertainty.
The configurations were chosen for this presentation after historical results
existed, not prospectively, and are not each task's selected maximum. Software
verification did not generate these estimates; they are not an independent replication.

\section{Dataset Description}
\label{sec:data-records}

The derivative dataset is publicly available from Zenodo at
\url{https://doi.org/10.5281/zenodo.21982978}
\cite{sanchis2026taskeegdata}. Its source inventories, QC evidence, continuous
signals, windows, and features are linked through shared pseudonymous
identifiers. Original recordings remain in BALLADEER
\cite{trujillo2026balladeerdata}.

\subsection{Coverage and analysis units}
\label{sec:data-overview}

The release contains 669 recordings from 156 participants, totalling 126.29
hours across five tasks and 35 task-family outputs.
Table~\ref{tab:data-overview} reports final task-level coverage.
Primary feature tables contain 107,478,560 records, with 6,308,526 additional
periodic-peak records. The table includes both types. Participant totals are
unique across tasks, whereas feature rows are repeated measurements and
structured outputs, not independent statistical observations.

\begin{table}[H]
\centering
\caption{Task-level derivative inventory.}
\label{tab:data-overview}
\small
\setlength{\tabcolsep}{3pt}
\begin{tabular}{@{}lrrrrr@{}}
\toprule
Task & Recordings & Participants & Channels & \shortstack{EEG\\hours} & \shortstack{Feature\\records} \\
\midrule
CogniFit & 129 & 129 & 14 & 80.89 & 58,004,337 \\
Robots & 123 & 123 & 14 & 10.02 & 7,099,948 \\
Slackline L1 & 140 & 140 & 29 & 11.88 & 16,365,912 \\
Slackline L6 & 139 & 139 & 29 & 11.80 & 16,226,365 \\
Slackline L11 & 138 & 138 & 29 & 11.71 & 16,090,524 \\
\midrule
Total & 669 & 156 & 14 or 29 & 126.29 & 113,787,086 \\
\bottomrule
\end{tabular}
\end{table}

\subsection{Repository and data organisation}
\label{sec:repository-release-layout}

The 54.61-GB dataset contains 48 downloadable files: 41 ZIP archives and seven
supporting files. Archives are organised by task and feature family, allowing
researchers to download only the components needed for their study. The main
file groups and formats are summarised in
Table~\ref{tab:repository-file-overview}.

\begin{table}[ht]
\centering
\caption{Dataset file groups and main formats.}
\label{tab:repository-file-overview}
\small
\begingroup
\setstretch{1.1}
\renewcommand{\arraystretch}{1.15}
\setlength{\tabcolsep}{3.5pt}
\begin{tabular}{@{}>{\raggedright\arraybackslash}p{0.26\textwidth}>{\raggedright\arraybackslash}p{0.27\textwidth}>{\raggedright\arraybackslash}p{0.39\textwidth}@{}}
\toprule
File group & Main formats & Contents \\
\midrule
Metadata and QC: one archive & CSV, Parquet and JSON & Participant and recording inventories, eligibility decisions, quality measurements and validation records \\
\addlinespace
Preprocessed EEG: five task-specific archives & MNE-FIF, with JSON, TSV and Parquet records & Continuous EEG signals, processing details and mappings between source and derivative recordings \\
\addlinespace
Features: 35 task-by-family archives & Parquet, with JSON and YAML records & Feature values, window and feature reference tables, calculation states and processing settings \\
\addlinespace
Seven supporting files & Markdown, CSV, JSON and plain text & Documentation, file inventory, release manifest, checksums, publication audit, version information and licence \\
\bottomrule
\end{tabular}
\endgroup
\end{table}

Shared participant, recording and window identifiers link these components.
The \emph{README.md} provides organisation and loading guidance;
\emph{ZENODO\_FILE\_INVENTORY.csv} lists each download's role, size and checksum.
Data dictionaries and feature catalogues explain fields, units, indexed values
and calculation states. Licence and processing records accompany the data.

For example, the
\emph{eegaware-eeg-derivatives-1.0.0-features-bal02-spectral.zip} archive
contains the conventional spectral features for Robots. Reference
tables included in each feature archive explain the feature definitions,
units, analysis windows, and ordering of stored values.

\subsection{Participant, task and label metadata}

Participant metadata contain one row per source participant, with pseudonym,
age, gender code, recruitment group, assessment status, and recording
availability. The fields \emph{group} and \emph{diagnosed} retain the meanings
in Section~\ref{subsec:source-cohort}. The category \emph{undetermined} is a
valid assessment outcome, not missing data.

Task-level records cover all expected participant-task pairs, including
missing recordings, and retain duration, sampling frequency, eligibility, and
decision reasons. Final selection records add QC outcomes, preprocessing plans,
and approved interpolation channels. The source field \emph{user\_id} and
later \emph{subject\_id} contain the same pseudonym. The \emph{recording\_id}
links participant-task records to signals, QC, and features. No names, contact
details, or dates of birth are included.

\subsection{Preprocessed EEG derivatives}

Signals retain 14 EPOC or 29 CGX channels in volts at 128 Hz, with template
coordinates. Table~\ref{tab:eeg-channel-schemas} specifies channel order and
excluded source fields. Interpolation preserves this schema and is recorded
separately.

\begin{table}[H]
\centering
\caption{EEG channel schemas in the released derivatives.}
\label{tab:eeg-channel-schemas}
\footnotesize
\setlength{\tabcolsep}{4pt}
\begin{tabular}{@{}>{\raggedright\arraybackslash}p{0.16\textwidth}>{\raggedright\arraybackslash}p{0.10\textwidth}>{\raggedright\arraybackslash}p{0.51\textwidth}>{\raggedright\arraybackslash}p{0.15\textwidth}@{}}
\toprule
Acquisition system & Tasks & Included scalp EEG channels, in order & Excluded channel-like fields \\
\midrule
EPOC X/Plus & CogniFit and Robots & AF3, F7, F3, FC5, T7, P7, O1, O2, P8, T8, FC6, F4, F8, AF4 & Device metrics and non-EEG fields \\
CGX Quick-32r & Slackline L1, L6 and L11 & AF7, Fpz, F7, Fz, T7, FC6, Fp1, F4, C4, Oz, CP6, Cz, PO8, CP5, O2, O1, P3, P4, P7, P8, Pz, PO7, T8, C3, Fp2, F3, F8, FC5, AF8 & A2, ExG 1, ExG 2, and non-EEG fields \\
\bottomrule
\end{tabular}
\end{table}

Each MNE-FIF signal file contains boundary annotations. Sample mappings link
its processed blocks to the source recording. Users can select recordings by
duration, block count and interpolation decisions from the cohort index before
loading the signals.

\subsection{QC and preprocessing records}
\label{subsec:qc-preprocessing-provenance-records}

Source audits describe each available recording's device, channels, timing,
and readability. Raw and postprocessing QC records contain temporal and
spectral measurements at the window, channel, and recording scales described
in Methods. Matched source and derivative measurements support comparisons
before and after preprocessing, with summaries of coverage and signal
consistency.

The accompanying documentation records acquisition strata, metric roles,
thresholds, and processing plans. Decision records retain automatic and final
QC outcomes, interpolation actions, and their reasons. Review assessments,
confidence, and rationale are preserved separately. The records
also document that clinical labels were excluded from processing decisions.

\subsection{Feature tables and catalogues}

Each task and feature family is accompanied by catalogues describing
recordings, channels, window configurations, analysis windows, and feature
definitions. The \emph{window\_id} and \emph{analysis\_interval\_id}
provide common temporal identifiers. Task and configuration information may
be encoded in folders rather than repeated in every table row. The supplied
loading example accounts for this organisation, and the file inventory
identifies the available task-family archives.

Feature-table rows differ in meaning and multiplicity across families
(Table~\ref{tab:feature-family-framework}). The edge catalogue fixes unique channel-pair ordering;
other catalogues define indexed values, including graph communities and
connectivity associations. These ordering rules are needed when unpacking or
joining outputs.

The feature catalogue specifies names, units, estimators, parameters,
applicability, and invalid-value rules. Calculation states distinguish
unsupported durations, absent peaks, undefined statistics, and failures.
These states must not be collapsed to zero or interpreted as equivalent
missingness mechanisms.

\subsection{Documentation and data integrity}
\label{subsec:documentation-integrity}

SHA-256 inventories verify downloaded archives and extracted files.
Dataset and software file manifests identify their contents, allowing the
resources used in a study to be documented precisely.

\section{Technical Validation Results}
\label{sec:technical-validation}

The following results describe source eligibility, curation, retained signal coverage, numerical reference checks and release integrity. Their interpretation is limited to the technical properties assessed.

\subsection{Source inventory and eligibility validation}
\label{subsec:source-inventory-validation}

\begin{figure}[htbp]
\centering
\includegraphics[width=\linewidth]{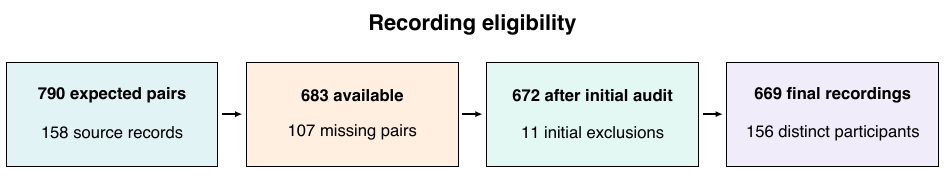}
\caption{Overall recording eligibility from expected pairs to final derivatives.}
\label{fig:recording-eligibility}
\end{figure}

Of 790 expected participant-task pairs across 158 participants, 107 lacked a
recording and 683 had a readable recording (Figure~\ref{fig:recording-eligibility}).
The initial audit excluded nine recordings for duration and two for sampling
frequency, leaving 672 (Table~\ref{tab:source-audit-summary} in
Appendix~\ref{app:supporting-numerical-tables}).

\begin{figure}[htbp]
\centering
\includegraphics[width=\linewidth]{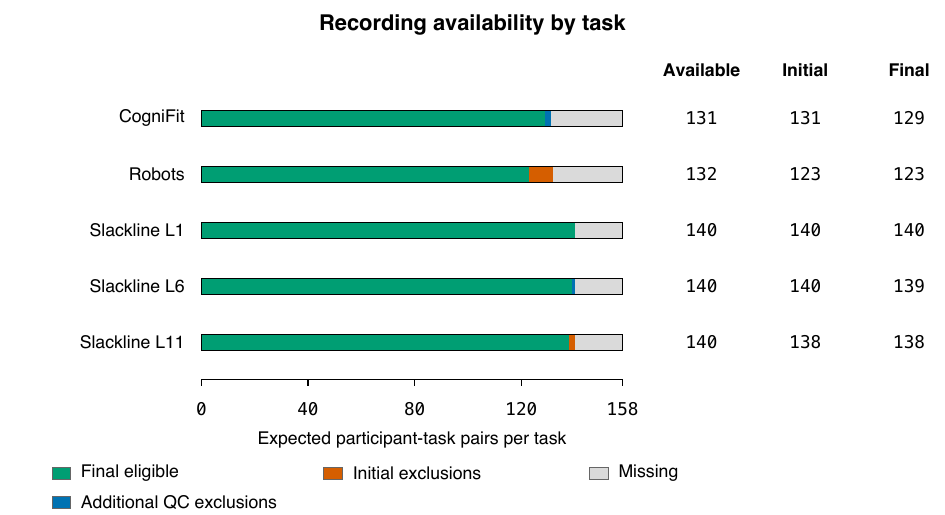}
\caption{Recording availability and retention across the five task conditions.}
\label{fig:task-availability}
\end{figure}

All 683 readable recordings completed raw QC without processing failure,
including those excluded by the initial audit. One of the four QC exclusions
had already failed the sampling-frequency check. The other three brought the
total to 14 distinct exclusions, leaving 669 eligible recordings from 156
participants. The QC records retain the measurements and decision reasons.
Eligibility was task-specific (Figure~\ref{fig:task-availability}); because
task cohorts overlap, their counts must not be summed as distinct participants.

\subsection{Raw QC and decision audit}
\label{subsec:raw-qc-decision-audit}

Label-blind calibration produced 222 stratum-specific thresholds, 346,309
window decisions, 15,862 channel decisions, and 683 recording decisions.
Continuous measurements and calculation states accompany the classifications.

The assessed subset was enriched for warnings and recordings requiring review.
These assessments supported curation and were not an error-free reference for
estimating false-positive rates, sensitivity, specificity, or cohort prevalence
\cite{delorme2023alone}.

Of the 683 readable recordings, 114 passed raw QC, 565 passed with warnings
and four were excluded by QC. A warning records an action or technical caveat;
it does not override the initial eligibility rules or exclude the participant.

\subsection{Preprocessing and post-processing QC validation}
\label{subsec:postprocessing-validation}

All 669 derivatives were generated without failure, with 217 channels
interpolated across 147 recordings.

In the paired raw and preprocessed 8s QC summaries, median slow-drift and
30--40 Hz ratios decreased in every task (Figure~\ref{fig:preprocessing-qc};
Table~\ref{tab:postprocessing-qc-summary} in Appendix~\ref{app:supporting-numerical-tables}).
Cross-channel spectral correlation remained similar for EPOC tasks and increased
for CGX tasks. It measures spectral structure across channels, not raw-to-processed
waveform fidelity. Connectors join task medians of recording-level measurements
at their tabulated precision, not individual paired observations. The figure
shows neither distributions nor uncertainty bars.

\begin{figure}[htbp]
\centering
\includegraphics[width=\linewidth]{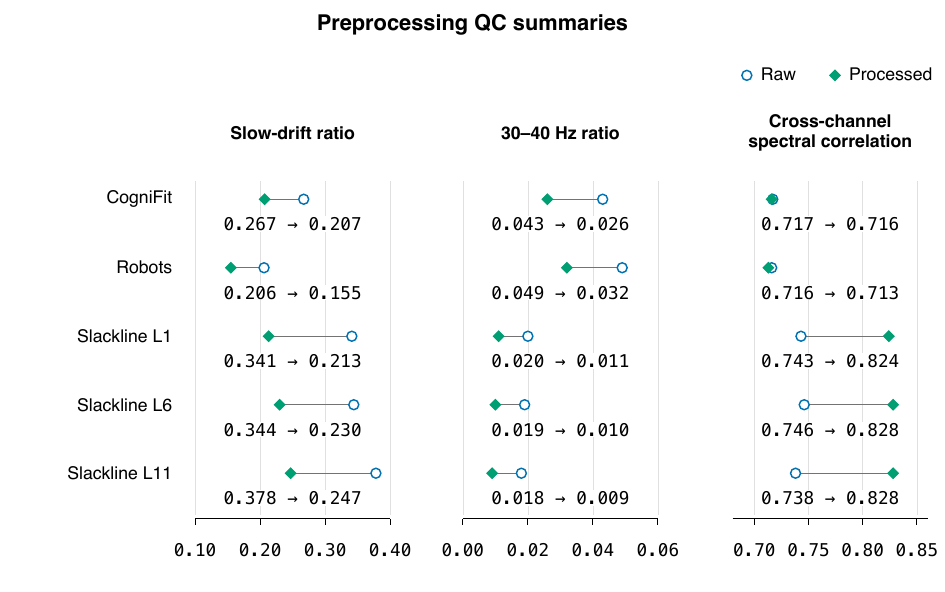}
\caption{Raw and processed recording-level QC medians by task.}
\label{fig:preprocessing-qc}
\end{figure}

Attenuation within targeted ranges, and above 40 Hz including 50 Hz, was
expected from the filter. These descriptive changes do not independently
establish improved neural-signal quality. Full paired values and distributions
remain available.

All 669 derivatives passed sample-count, 128-Hz sampling, block-correspondence,
finite-sample, and channel-inventory checks. Median retained duration was
100\% in every task. Two short EPOC blocks totalling 0.211 s were omitted,
with a minimum retained proportion of 99.978\%. Declared and observed block
counts agreed throughout.

\subsection{Derivative integrity and computational reproducibility}
\label{subsec:derivative-integrity-validation}

The integrity audit confirmed readable derivatives with expected annotations,
continuous-block records and metadata. Original recordings and preceding QC records remain unchanged.
Download verification is described in Section~\ref{subsec:documentation-integrity}.

\subsection{Feature release validation}
\label{subsec:feature-release-validation}

Every family recorded \emph{validated} status.
Table~\ref{tab:feature-numerical-validation} gives the saved outcomes and
their bounded interpretation. These are executed numerical checks, not merely
test-source availability, and do not establish biological validity or adequate
performance on every real epoch.

\begin{table}[H]
\centering
\caption{Saved numerical reference checks by feature family.}
\label{tab:feature-numerical-validation}
\small
\begingroup
\setstretch{1.05}
\renewcommand{\arraystretch}{1.18}
\setlength{\tabcolsep}{4pt}
\begin{tabular}{@{}>{\raggedright\arraybackslash}p{0.19\textwidth}>{\raggedright\arraybackslash}p{0.43\textwidth}>{\raggedright\arraybackslash}p{0.28\textwidth}@{}}
\toprule
Family & Reference checks & Recorded outcome and scope \\
\midrule
Temporal and statistical & Constant mean and variance, sinusoidal RMS, and alternating-signal zero-crossing rate. & Listed checks passed for the specified signals. \\
Conventional spectral & Known sinusoidal bands and dominant frequencies, zero power for a constant signal, and frequency-grid spacing. & Expected values and grid checks passed. \\
Aperiodic spectral & Pure-spectrum offset and exponent recovery, near-zero residuals, and known theta and alpha peaks. & Parameter and peak-recovery checks passed for synthetic spectra. \\
Complexity and fractal & Constant and ramp limits, sinusoid-versus-noise ordering, and persistence ordering. & Listed limits and orderings passed, with explicit undefined values. \\
Functional connectivity & Identical, inverted, and constant signals tested signed correlation and spectral limiting cases. & Expected values and undefined constant-signal correlation were confirmed. \\
Graph and network & Complete, path, star, and disconnected graphs tested topology, sign decomposition, and Laplacian invariants. & Listed graph properties and invariants passed. \\
Time-frequency decomposition & Wavelet reconstruction and percentage-sum invariants, with saved reference cases. & Invariants recorded as validated per row. \\
\bottomrule
\end{tabular}
\endgroup
\end{table}

A separate integrity audit confirmed all 35 task-family outputs and 1,832
files, with no identified integrity issues. Recording and window inventories,
feature definitions, validation records, and checksums were present. All
families were linked to the same preprocessed source recordings, and graph
outputs were linked to their corresponding connectivity values.

Temporal/statistical, spectral, and decomposition outputs had no invalid
scalar values. Aperiodic records distinguish absent accepted peaks from fit
warnings or failures. The accompanying fit diagnostics retain
calculation states, warning reasons, squared correlation, and MAE, with summaries
by task, device, and epoch configuration. These empirical diagnostics should
guide selection for reuse. Finite parameters and successful synthetic checks
do not guarantee adequate fits to all recorded spectra.

Complexity outputs retain four duration-inapplicable nulls at 2s and one at
4s. Connectivity and graph outputs retain mathematically undefined states.
Unsupported duration, absent peak, and undefined statistic must not be pooled
as one processing-failure category. Catalogues distinguish definition counts,
indexed positions, and physical rows. Together, the audits support numerical
properties, structural completeness, and traceable lineage, without using
classification performance as evidence of clinical validity.

\section{Discussion}
\label{sec:discussion}

Reuse shifts effort from reconstructing signal preparation and window alignment
to choosing scientifically appropriate subsets and representations. Eligibility,
interpolation and calculation-state records make those choices inspectable,
while the retained continuous signals allow researchers to test alternative
estimators against the supplied measurements.

\subsection{Predictive utility demonstration}
\label{subsec:predictive-utility-results}

The saved evaluations illustrate predictive reuse separately from technical
validation (Figure~\ref{fig:predictive-utility}; Table~\ref{tab:predictive-utility}
in Appendix~\ref{app:supporting-numerical-tables}). AUROC describes
recruitment-group discrimination; lower Brier scores indicate smaller probability
error, not complete calibration. The dashed AUROC 0.5 line is a reference, not
the comparator's measured result. No Brier intervals or significance tests are
shown. Retrospective selection and fixed-prediction uncertainty are described in
Section~\ref{subsec:predictive-utility-methods}.

\begin{figure}[htbp]
\centering
\includegraphics[width=\linewidth]{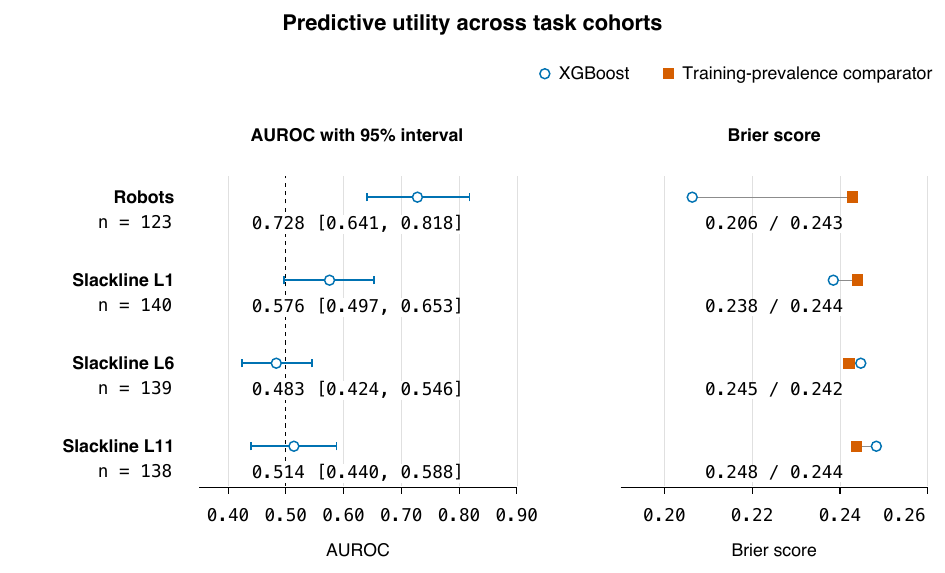}
\caption{Saved predictive utility across four overlapping task cohorts.}
\label{fig:predictive-utility}
\end{figure}
\FloatBarrier

Robots had the highest AUROC point estimate, whereas Slackline discrimination
was weaker. XGBoost's Brier score was lower than the prevalence comparator's
for Robots and L1, but higher for L6 and L11. These descriptive differences
are not superiority tests. Overlapping cohorts, acquisition differences and
task-specific partitions prevent treating tasks as independent cohorts or
interpreting their differences causally.

\subsection{Research applications and processing trade-offs}

Researchers can investigate associations between task EEG and participant characteristics, compare complementary feature representations, and examine feature redundancy or selection stability. These applications require explicit analytical targets and participant-aware evaluation.

The preprocessing sequence balances comparability with recording coverage.
Its effects under filtering, interpolation and average reference must be judged
against the intended analysis and residual contamination, rather than attenuation
alone. Numerical reference checks and release audits establish estimator
properties and structural completeness; biological interpretation requires
additional evidence.

\subsection{Limitations and appropriate interpretation}
\label{subsec:processing-limitations}

Transient artefacts and artefact components were not removed. Residual ocular,
muscle, and movement contamination can remain. Interpolation and average
reference can affect connectivity and graphs, while common filtering and
sampling cannot remove montage differences or task-device confounding.

Recruitment \emph{group} and assessment \emph{diagnosed} are distinct.
Do not reassign \emph{undetermined} to a diagnostic class without an explicit
external rule. Within-task analyses simplify interpretation because task,
device, montage, and duration differ. Pooling cannot separate task and device
effects, and common sampling does not establish hardware equivalence.
Multimodal linkage requires checking task-specific timing, not just identifiers.

The source cohort and task-acquisition design bound generalisation. Recruitment and assessment labels need to be interpreted in their documented clinical context. The resource supports methodological and neurophysiological research, while clinical or diagnostic applications require independent validation. Aperiodic fits and mathematically undefined outputs require review of their saved diagnostics and calculation states before analytical inclusion.

\subsection{Practical reuse guidance}
\label{sec:usage-notes}

Begin with \emph{README.md}, use the file inventory to select the required
archives, and verify downloads against \emph{SHA256SUMS}. Signals can be read
with MNE-FIF-compatible tools, and tables with CSV/Parquet readers. Retain the
accompanying calculation states and quality records when selecting data.

A manageable starting point is the conventional spectral feature set for
Robots (BAL02), using 8s windows. The supplied loading example shows how to select this
task, family, and duration. Consult the feature and window catalogues to
identify the measurements and their sample boundaries. Match channel-level
families using recording, window, channel schema, and channel identifiers.
Matching only by window would incorrectly combine measurements from different
channels. Use participant identifiers to link metadata and the family
catalogues to interpret indexed values, rather than relying on row order.

For predictive evaluation, keep all recordings and epochs from each
participant in one training, validation, or test partition. Learn imputation,
scaling, feature selection, dimensionality reduction, and model settings from
training data only, using inner validation where required. Interpret nulls
through their recorded reasons rather than replacing them with zero.

Selecting only the required tasks, feature families, durations, and columns
reduces memory requirements. Researchers evaluating different preprocessing
should begin with the original BALLADEER signals and document their processing.

\section{Conclusions}

Differences between EEG features are difficult to interpret when the features
come from different stretches of a recording or from signals prepared in
different ways. To support matched comparisons in ADHD research, we curated
669 task recordings from 156 children and adolescents in BALLADEER and
linked seven families of features to the continuous EEG from which they were
calculated. Neither recruitment group nor diagnostic assessment determined
which recordings were retained. Researchers can compare, for
example, spectral power and connectivity over the same eight seconds of a
task, or compute a new measure on those samples. The accompanying records
identify interpolated channels and distinguish a window too short for an
estimator from one in which no spectral peak met the acceptance criteria.

All 669 preprocessed recordings passed checks on sample counts, channel inventories and
continuity boundaries. Each feature family also passed its numerical
reference checks. These results document which samples were retained and how
the estimators behaved on known inputs; they do not rule out residual movement
or muscle artefacts.

The saved predictive example covered four of five conditions: Robots, a visual search game,
and three Slackline levels in which players respond to on-screen flags.
Using eight-second windows to distinguish recruitment groups in held-out
participants, XGBoost yielded a mean area under the receiver operating characteristic curve of 0.728 for
Robots and 0.483--0.576 across Slackline levels. The cohorts overlap and
recording systems differ, so the results neither isolate a task effect nor validate an
ADHD diagnostic test. Within a task, researchers can test whether connectivity
adds information beyond spectral power using the same windows and held-out
participants.

\clearpage
\section*{Data Availability}

The dataset (version 1.0.0, CC BY 4.0) is publicly available from Zenodo at
\url{https://doi.org/10.5281/zenodo.21982978} \cite{sanchis2026taskeegdata}.
The original BALLADEER dataset is available from Figshare at
\url{https://doi.org/10.6084/m9.figshare.28676042.v2}
\cite{trujillo2026balladeerdata}.

\section*{Code Availability}

The curation software, code for the XGBoost example and saved aggregate results
are available together in version 1.1.0 on Zenodo at
\url{https://doi.org/10.5281/zenodo.24399993} \cite{sanchis2026eegsoftware110}.
The software is licensed under BSD 3-Clause; the saved aggregate results and new
release documentation are licensed under CC BY 4.0. The original curation release
(version 1.0.0) remains available at
\url{https://doi.org/10.5281/zenodo.21982948} \cite{sanchis2026eegsoftware}.

\section*{CRediT authorship contribution statement}

Author contributions are reported using the Contributor Roles Taxonomy (CRediT).
\textbf{Javier Sanchis}: Conceptualization, Data curation, Formal analysis,
Investigation, Methodology, Resources, Software, Supervision, Validation,
Visualization, Writing - original draft, Writing - review \& editing.
\textbf{{\'A}lvaro Navarro} and \textbf{Jorge Garc{\'i}a-Carrasco}:
Conceptualization, Data curation, Formal analysis, Investigation, Methodology,
Resources, Validation, Visualization, Writing - original draft,
Writing - review \& editing.
\textbf{Tahar Kechadi} and \textbf{Juan Trujillo}: Conceptualization,
Formal analysis, Funding acquisition, Project administration, Supervision,
Writing - review \& editing.

\section*{Funding}

This work was supported by Subvenciones para estancias de personal investigador
doctor en centros de investigaci{\'o}n radicados fuera de la Comunitat Valenciana
para el ejercicio 2026 (CIBEST-2026), grant CIBEST/2026/181, Conselleria
d'Educaci{\'o}, Cultura i Universitats. It was also supported by the BALIDA-AA
project (CIPROM/2024/13), funded by Conselleria de Educaci{\'o}n, Cultura,
Universidades y Empleo (Generalitat Valenciana). Additional support was
provided by the KOSMOS-UA project (PID2024-155363OB-C43), funded by the Spanish
Ministry of Science and Innovation. The funders had no role in study design,
analysis, writing or the decision to submit.

\section*{Competing Interests}

The authors declare no competing interests.

\section*{Declaration of generative AI and AI-assisted technologies in the manuscript preparation process}

During the preparation of this work, the authors used ChatGPT and Codex to
assist manuscript review, evidence checking, drafting, rewriting and sentence
polishing. After using these tools, the authors reviewed and edited the content
as needed and take full responsibility for the content of this manuscript.

\clearpage
\PreprintBibliography{references}
\clearpage
\appendix
\numberwithin{equation}{section}
\renewcommand{\theequation}{\Alph{section}.\arabic{equation}}
\renewcommand{\thesubsection}{\Alph{section}.\arabic{subsection}}
\setcounter{table}{0}
\renewcommand{\thetable}{\Alph{section}.\arabic{table}}
\section{Mathematical feature definitions}
\label{app:feature-definitions}

This appendix gives the equations and fixed estimator conventions underlying
the feature inventories in Section~\ref{subsec:feature-framework}. The
definitions apply independently to each valid gap-aware epoch and retain the
units of the preprocessed EEG signal. Undefined mathematical states were stored
as null values with an explicit reason rather than replaced by zero.

\subsection{Temporal and statistical definitions}
\label{app:temporal-statistical-definitions}

For channel epoch $x=(x_1,\ldots,x_N)$ sampled at $f_s$, mean and mean absolute
value use denominator $N$. Variance $s_x^2$ uses $N-1$ and represents Hjorth activity. SD is $s_x$, and
RMS is $\sqrt{N^{-1}\sum_i x_i^2}$. Other summaries are minimum, maximum,
median, peak-to-peak range, IQR with linearly interpolated quantiles, and
unscaled MAD about the median. For central moments
$m_r=N^{-1}\sum_i(x_i-\bar{x})^r$, adjusted skewness and unbiased excess
kurtosis are
\begin{equation}
\begin{aligned}
G_1&=\frac{\sqrt{N(N-1)}}{N-2}\frac{m_3}{m_2^{3/2}},\\
G_2&=\frac{N-1}{(N-2)(N-3)}
\left[(N+1)\left(\frac{m_4}{m_2^2}-3\right)+6\right].
\end{aligned}
\end{equation}
Insufficient sample size or zero variance gives a null. Line-length and
zero-crossing rates are
\begin{equation}
\operatorname{LLR}=\frac{f_s}{N-1}\sum_{i=2}^N|x_i-x_{i-1}|,\qquad
\operatorname{ZCR}=\frac{f_s}{N-1}C(x),
\end{equation}
where $C(x)$ counts sign transitions \cite{esteller2001linelength}. Values
below double-precision machine epsilon are treated as zero, with internal
zero runs inheriting the preceding non-zero sign.

With $d_i=f_s(x_i-x_{i-1})$, $e_i=f_s(d_i-d_{i-1})$, and sample variance
$\operatorname{Var}_s$, Hjorth measures \cite{hjorth1970eeg} are
\begin{equation}
\operatorname{Mobility}=\sqrt{\frac{\operatorname{Var}_s(d)}{\operatorname{Var}_s(x)}},
\qquad
\operatorname{Complexity}=\frac{\sqrt{\operatorname{Var}_s(e)/\operatorname{Var}_s(d)}}
{\operatorname{Mobility}}.
\end{equation}
Zero denominators produce reason-coded nulls.

\subsection{Conventional spectral definitions}
\label{app:spectral-definitions}

The one-sided Welch PSD $S_x(f)$ uses 2s periodic Hann segments, 50\% overlap,
constant detrending, and 256 Fourier points at 128 Hz: 256 samples per segment,
128 overlapping samples, and a 0.5-Hz grid \cite{welch1967fft,harris1978windows}.
Exact band boundaries are inserted by linear interpolation before trapezoidal
integration. Bands are delta $[1,4]$, theta $[4,8]$, alpha $[8,13]$, beta
$[13,30]$, and restricted gamma $[30,40]$ Hz. Absolute power is
$P_B=\int_B S_x(f)\,df$, total power is $P_{1\text{-}40}$, and relative power
is $P_B/P_{1\text{-}40}$. Ratios are $P_\theta/P_\beta$, $P_\theta/P_\alpha$,
and $P_\delta/P_\gamma$, with nulls for zero denominators.

Dominant frequency is the lowest maximising bin within 1--40 Hz. For its $M$
bins, $p_j=S_x(f_j)/\sum_k S_x(f_k)$ and normalised spectral entropy is
$-\sum_jp_j\ln p_j/\ln M$ \cite{inouye1991spectralentropy}. Spectral edges
at 50\% and 95\% locate the first cumulative-power interval reaching the
corresponding fraction of $P_{1\text{-}40}$, with linear interpolation of the
crossing frequency.

\subsection{Aperiodic spectral definitions}
\label{app:aperiodic-definitions}

The 1--30 Hz log-power spectrum $Y(f)=\log_{10}S_x(f)$ uses fixed no-knee
background $L(f)$ plus Gaussian peaks
\cite{donoghue2020parameterizing,gerster2022separating}:
\begin{equation}
L(f)=b-\chi\log_{10}f,\qquad
\widehat{Y}(f)=L(f)+\sum_{n=1}^{K}a_n
\exp\!\left[-\frac{(f-c_n)^2}{2w_n^2}\right].
\end{equation}
Here $b$ is offset, $\chi$ the positive exponent, and each peak has height
$a_n$, centre $c_n$, and bandwidth $2w_n$. Peak widths are 1--8 Hz, minimum
height 0.1 log-power units, threshold 2.5 residual SD, and maximum count six.

The highest-power accepted peak in theta $[4,8)$, alpha $[8,13)$, and beta
$[13,30]$ supplies centre, power, and bandwidth. Ties select the lower centre,
then narrower bandwidth. No accepted peak gives a null triplet. The 16
predictors comprise offset, exponent, peak count, nine band-peak values, and
four adjusted band means. For delta, theta, alpha, and beta bounds $[l,u]$,
the latter are $(u-l)^{-1}\int_l^u[Y(f)-L(f)]\,df$.
Fit diagnostics are squared correlation $\operatorname{corr}(Y,\widehat{Y})^2$
and MAE $M^{-1}\sum_j|Y(f_j)-\widehat{Y}(f_j)|$. They are excluded from the
default predictor view.

\subsection{Complexity and fractal definitions}
\label{app:complexity-definitions}

All estimators use channel epochs at 128 Hz. Higuchi, Katz, Petrosian,
permutation, dispersion, singular-value, and LZ76 descriptors support 2s.
Rescaled-range and detrended fluctuation exponents and sample entropy require
4s, and approximate entropy requires 8s.

Higuchi dimension \cite{higuchi1988fractal} is the least-squares slope of
$\log L(k)$ against $\log(1/k)$ for $k=1,\ldots,10$, with
\begin{equation}
\begin{aligned}
N_m&=\left\lfloor\frac{N-m}{k}\right\rfloor,\\
L_m(k)&=\frac{N-1}{k^2N_m}\sum_{j=1}^{N_m}|x_{m+jk}-x_{m+(j-1)k}|,\\
L(k)&=\frac1k\sum_{m=1}^k L_m(k).
\end{aligned}
\end{equation}
Katz dimension \cite{katz1988fractals} is
$D_K=\log_{10}(L/a)/\log_{10}(d/a)$, where
$L=\sum_{i=1}^{N-1}|x_{i+1}-x_i|$, $a=L/(N-1)$, and
$d=\max_i|x_i-x_1|$. Zero path, mean step, diameter, or denominator gives a
null. Petrosian dimension \cite{petrosian1995complexity} uses the number
$N_\Delta$ of first-difference sign changes:
\begin{equation}
D_P=\frac{\log_{10}N}{\log_{10}N+\log_{10}[N/(N+0.4N_\Delta)]}.
\end{equation}

Rescaled-range (R/S) analysis \cite{hurst1951storage} uses complete blocks of
$n\in\{16,32,64,128\}$ samples. Within each block, $R$ is the range of the
cumulative demeaned signal and $S$ its original sample SD. The Hurst estimate
is the least-squares slope of $\log\operatorname{mean}(R/S)$ against
$\log n$, using valid block ratios. Detrended Fluctuation Analysis (DFA)
\cite{peng1994dfa} first integrates the demeaned epoch, removes a linear
trend within each complete block at the same scales, and fits the slope of
$\log F(n)$ against $\log n$, where $F(n)$ is the RMS residual across all
samples in those blocks.

Approximate and sample entropy \cite{pincus1991approxentropy,richman2000sampleentropy}
use dimension $m=2$, delay one sample, Chebyshev distance, and tolerance
$r=0.2\sigma_x$ with population SD $\sigma_x$. For templates
$u_i^{(m)}=(x_i,\ldots,x_{i+m-1})$, define
\begin{equation}
C_i^{(m)}=\frac{\#\{j:\|u_i^{(m)}-u_j^{(m)}\|_\infty\le r\}}{N-m+1},
\qquad \phi_m=\frac1{N-m+1}\sum_i\ln C_i^{(m)}.
\end{equation}
Approximate entropy is $\phi_m-\phi_{m+1}$, including self-matches. Sample
entropy is $-\ln(A/B)$ for distinct matching pairs of lengths $m+1$ and $m$,
respectively. Insufficient matches give a null.

\begin{sloppypar}
Permutation entropy \cite{bandt2002permutation} uses order three, delay one,
and $-\sum_\pi p(\pi)\ln p(\pi)/\ln6$. Dispersion entropy
\cite{rostaghi2016dispersion} maps samples to
$y_i=\Phi[(x_i-\bar{x})/\sigma_x]$ and classes
$z_i=\operatorname{clip}(\lfloor6y_i+1\rfloor,1,6)$, then uses the same
entropy form divided by $\ln36$ for length-two class patterns. Singular-value
entropy uses a centred order-three, delay-one embedding, singular-value
proportions $q_j=\sigma_j/\sum_\ell\sigma_\ell$, and
$-\sum_{j=1}^3q_j\log_2q_j/\log_2 3$.
LZ76 \cite{lempel1976complexity} encodes samples at or above the median as one
and others as zero, counts new phrases by left-to-right parsing, and
normalises the count as $C_{\mathrm{LZ}}\log_2N/N$.
\end{sloppypar}

\subsection{Functional connectivity definitions}
\label{app:connectivity-definitions}

Broadband Pearson correlation retains its sign, without lag search or Fisher
transformation. Spectral estimates use the 2s periodic Hann geometry in
Appendix~\ref{app:spectral-definitions}. Let $Q_{ij,k}(f)$ be segment $k$'s
cross-periodogram, $\widehat S_{ij}$ its mean over $K$ segments, and
$\operatorname{Coh}_{ij}=\widehat S_{ij}/\sqrt{\widehat S_{ii}\widehat S_{jj}}$.
Magnitude-Squared Coherence (MSC) and absolute imaginary coherency are
\begin{equation}
\operatorname{MSC}_{ij}=|\operatorname{Coh}_{ij}|^2,\qquad
\operatorname{ImC}^{\mathrm{abs}}_{ij}=|\operatorname{Im}(\operatorname{Coh}_{ij})|.
\end{equation}
Imaginary coherency reduces instantaneous field-spread effects but does not
remove all sensor-space confounds \cite{nolte2004imaginary,bastos2016connectivity}.

For non-zero cross-periodograms, let $U_{ij,k}=Q_{ij,k}/|Q_{ij,k}|$.
Phase-Locking Value (PLV), Phase-Lag Index (PLI), and weighted Phase-Lag Index
(wPLI) \cite{lachaux1999phase,stam2007pli,vinck2011wpli} are
\begin{align}
\operatorname{PLV}_{ij}&=\left|K^{-1}\sum_kU_{ij,k}\right|, &
\operatorname{PLI}_{ij}&=\left|K^{-1}\sum_k\operatorname{sign}(\operatorname{Im}Q_{ij,k})\right|,\\
\operatorname{wPLI}_{ij}&=\frac{|\sum_k\operatorname{Im}Q_{ij,k}|}
{\sum_k|\operatorname{Im}Q_{ij,k}|}. &&
\end{align}
A zero wPLI denominator gives a null. Estimates are averaged across discrete
bins in the five spectral bands. Matrices are symmetric with unit diagonal and
canonical channel-pair order. Pearson supports all durations, while spectral
matrices require at least seven segments and therefore 8s/16s epochs.

\subsection{Graph and network definitions}
\label{app:graph-definitions}

Non-negative adjacency matrices $W$ use $|r|$, $\max(r,0)$, and $\max(-r,0)$
for Pearson correlation and direct weights for 25 spectral matrices. Positive
weights define edges of length $1/W_{ij}$, while zero denotes an absent edge
with infinite length \cite{rubinov2010network}. For $C$ channels, define
strength $s_i=\sum_jW_{ij}$, degree $k_i$, neighbours $V_i$, and
$\widehat W=W/\max_{u<v}W_{uv}$.

Normalised node strength is $s_i/(C-1)$. Onnela clustering
\cite{onnela2005weighted} and modified weighted local efficiency
\cite{latora2001efficiency,rubinov2010network} are
\begin{align}
C_i&=\frac{\sum_{j\ne h\in V_i}
(\widehat W_{ij}\widehat W_{ih}\widehat W_{jh})^{1/3}}{k_i(k_i-1)},\\
E_i^{\mathrm{loc}}&=\frac{\sum_{j\ne h\in V_i}
\widehat W_{ij}^{1/3}\widehat W_{ih}^{1/3}[d_{jh}^{(i,1/3)}]^{-1}}{k_i(k_i-1)}.
\end{align}
The sums run over ordered distinct neighbours. Distances $d_{jh}^{(i,1/3)}$
are shortest paths within that neighbour subgraph using lengths
$\widehat W^{-1/3}$. Both metrics are zero for $k_i<2$.
Normalised betweenness is
$2[(C-1)(C-2)]^{-1}\sum_{s<t,\,s,t\ne i}\sigma_{st}(i)/\sigma_{st}$,
where $\sigma_{st}(i)$ counts weighted shortest paths through $i$ and
$\sigma_{st}$ all such paths. Eigenvector centrality is the unit-length
principal eigenvector of $W$, with non-negative sum.

Global summaries include mean weight over all possible undirected edges,
strength coefficient of variation using population SD, mean clustering, and
mean local efficiency. With shortest distances $d_{ij}$ under $1/W$,
\begin{equation}
E^{\mathrm{global}}=\frac{\sum_{i\ne j}d_{ij}^{-1}}{C(C-1)},\qquad
L=\frac{\sum_{i\ne j}d_{ij}}{C(C-1)}.
\end{equation}
Unreachable pairs contribute zero efficiency, but make characteristic path
length $L$ null. Weighted modularity \cite{newman2004modularity} is
\begin{equation}
Q=\frac1{2m}\sum_{i,j}\left(W_{ij}-\frac{s_is_j}{2m}\right)\delta(c_i,c_j),
\qquad m=\sum_{i<j}W_{ij},
\end{equation}
using deterministic greedy communities $c_i$ at resolution one. For the
Laplacian $\operatorname{diag}(s)-W$ with eigenvalues
$0=\lambda_1\le\lambda_2\le\cdots\le\lambda_C$, algebraic connectivity is
$\lambda_2$, spectral radius is $\lambda_C$, and synchronisability is
$\lambda_C/\lambda_2$, null when $\lambda_2$ is numerically zero.

\subsection{Time-frequency decomposition definitions}
\label{app:decomposition-definitions}

Four-level DWTs use \emph{db4}, \emph{coif2}, and \emph{sym5} with
periodisation boundaries \cite{mallat1989wavelet,daubechies1988wavelets}.
Ordered coefficients $[cA_4,cD_4,cD_3,cD_2,cD_1]$ represent nominal
0--4, 4--8, 8--16, 16--32, and 32--64 Hz intervals. The final carrier
contains available 32--40 Hz activity and transition effects, not restored
information above the preprocessing cutoff.

Let $s_b[n]$ reconstruct carrier $b$ with all other coefficient sets zero,
and $E_b=\sum_{n=0}^{N-1}s_b[n]^2$. Carrier summaries are mean power $E_b/N$,
mean absolute amplitude, adjusted skewness, and unbiased excess kurtosis as in
Appendix~\ref{app:temporal-statistical-definitions}. With
$p_{b,n}=s_b[n]^2/E_b$, energy entropy is
$-\sum_np_{b,n}\ln p_{b,n}/\ln N$. Mean Teager--Kaiser energy
\cite{kaiser1990energy} is
\begin{equation}
\overline\Psi_b=\frac1{N-2}\sum_{n=1}^{N-2}
\big[s_b[n]^2-s_b[n-1]s_b[n+1]\big].
\end{equation}
For five carriers $B$, Percentage Raw Energy (PRE) is $100E_b/\sum_{r\in B}E_r$.
Wavelet energy entropy is $-\sum_bq_b\ln q_b/\ln5$, where
$q_b=E_b/\sum_rE_r$.

Amplitude modulation uses analytic envelopes
$e_b[n]=|s_b[n]+j\mathcal H\{s_b[n]\}|$, with discrete Hilbert transform
$\mathcal H$ \cite{safi2026amplitude}. For
$V_b=\sum_n(e_b[n]-\bar e_b)^2$, Percentage Envelope Energy (PEE) is
$100V_b/\sum_rV_r$. Each envelope undergoes the same four-level DWT, retaining
modulation intervals 0--4, 4--8, 8--16, and 16--32 Hz. With reconstructed
modulation $m_{b,r}[n]$ and energy $M_{b,r}=\sum_nm_{b,r}[n]^2$, Percentage
Modulation Energy (PME) is $100M_{b,r}/\sum_{u\in B}\sum_{v\in R}M_{u,v}$.
Ten templates yield 61 values per wavelet and 183 across the three wavelets.

\clearpage
\section{Supporting numerical tables}
\label{app:supporting-numerical-tables}
\setcounter{table}{0}

The following tables retain the task-level initial-audit exclusions,
preprocessing QC medians and saved predictive-utility estimates accompanying
Figures~\ref{fig:task-availability}, \ref{fig:preprocessing-qc} and
\ref{fig:predictive-utility}. Initial-audit retention precedes raw-signal QC
and is distinct from final eligibility. The predictive-utility table also
reports balanced accuracy, which is not plotted.

\begin{table}[ht]
\centering
\caption{Source recording availability and initial-audit eligibility by task.}
\label{tab:source-audit-summary}
\small
\setlength{\tabcolsep}{2pt}
\begin{tabular}{@{}>{\raggedright\arraybackslash}p{0.23\textwidth}rrrrrr@{}}
\toprule
Task & \shortstack{Cohort\\subjects} & \shortstack{Available\\recordings} &
\shortstack{Missing\\recordings} & \shortstack{Duration\\exclusions} &
\shortstack{Sampling\\frequency\\exclusions} & \shortstack{Retained after\\initial audit} \\
\midrule
CogniFit & 158 & 131 & 27 & 0 & 0 & 131 \\
Robots & 158 & 132 & 26 & 7 & 2 & 123 \\
Slackline L1 & 158 & 140 & 18 & 0 & 0 & 140 \\
Slackline L6 & 158 & 140 & 18 & 0 & 0 & 140 \\
Slackline L11 & 158 & 140 & 18 & 2 & 0 & 138 \\
\midrule
Total & & 683 & 107 & 9 & 2 & 672 \\
\bottomrule
\end{tabular}
\end{table}

\FloatBarrier
\begin{table}[H]
\centering
\caption{Paired recording-level QC medians before and after preprocessing.}
\label{tab:postprocessing-qc-summary}
\small
\setlength{\tabcolsep}{3pt}
\begin{tabular}{@{}lrrrr@{}}
\toprule
Task & Recordings & \shortstack{Slow-drift\\ratio} & \shortstack{30-40 Hz\\ratio} & \shortstack{Spectral\\correlation} \\
\midrule
CogniFit & 129 & $0.267 \rightarrow 0.207$ & $0.043 \rightarrow 0.026$ & $0.717 \rightarrow 0.716$ \\
Robots & 123 & $0.206 \rightarrow 0.155$ & $0.049 \rightarrow 0.032$ & $0.716 \rightarrow 0.713$ \\
Slackline L1 & 140 & $0.341 \rightarrow 0.213$ & $0.020 \rightarrow 0.011$ & $0.743 \rightarrow 0.824$ \\
Slackline L6 & 139 & $0.344 \rightarrow 0.230$ & $0.019 \rightarrow 0.010$ & $0.746 \rightarrow 0.828$ \\
Slackline L11 & 138 & $0.378 \rightarrow 0.247$ & $0.018 \rightarrow 0.009$ & $0.738 \rightarrow 0.828$ \\
\bottomrule
\end{tabular}
\end{table}

\FloatBarrier
\begin{table}[htbp]
\centering
\small
\caption{Participant-level recruitment-group discrimination from saved ALL/XGBoost/8s evaluations.}
\label{tab:predictive-utility}
\setlength{\tabcolsep}{4pt}
\begin{tabularx}{\linewidth}{@{}>{\raggedright\arraybackslash}X r c c c c@{}}
\toprule
Task & $n$ & \shortstack{XGBoost AUROC\\(95\% CI)} & \shortstack{XGBoost\\balanced\\accuracy} & \shortstack{XGBoost\\Brier score} & \shortstack{Prevalence\\comparator\\Brier score} \\
\midrule
Robots & 123 & 0.728 (0.641--0.818) & 0.652 & 0.206 & 0.243 \\
Slackline L1 & 140 & 0.576 (0.497--0.653) & 0.517 & 0.238 & 0.244 \\
Slackline L6 & 139 & 0.483 (0.424--0.546) & 0.497 & 0.245 & 0.242 \\
Slackline L11 & 138 & 0.514 (0.440--0.588) & 0.487 & 0.248 & 0.244 \\
\bottomrule
\end{tabularx}
\par\smallskip
\begin{minipage}{\linewidth}
\small Lower Brier scores are better. The prevalence comparator's stored balanced
accuracy is 0.500 in all four cohorts. Counts represent overlapping participant
cohorts, not independent samples across tasks. Values are rounded directly from
stored configuration summaries distributed with software version 1.1.0
\cite{sanchis2026eegsoftware110}.
\end{minipage}
\end{table}

\FloatBarrier

\end{document}